# SwarmTouch: Tactile Interaction of Human with Impedance Controlled Swarm of Nano-Quadrotors

E. Tsykunov, L. Labazanova, A. Tleugazy, D. Tsetserukou

*Abstract*— We propose a novel interaction strategy for a human-swarm communication when a human operator guides a formation of quadrotors with impedance control and receives vibrotactile feedback. The presented approach takes into account the human hand velocity and changes the formation shape and dynamics accordingly using impedance interlinks simulated between quadrotors, which helps to achieve a life-like swarm behavior. Experimental results with Crazyflie 2.0 quadrotor platform validate the proposed control algorithm. The tactile patterns representing dynamics of the swarm (extension or contraction) are proposed. The user feels the state of the swarm at his fingertips and receives valuable information to improve the controllability of the complex life-like formation. The user study revealed the patterns with high recognition rates. Subjects stated that tactile sensation improves the ability to guide the drone formation and makes the human-swarm communication much more interactive. The proposed technology can potentially have a strong impact on the human-swarm interaction, providing a new level of intuitiveness and immersion into the swarm navigation.

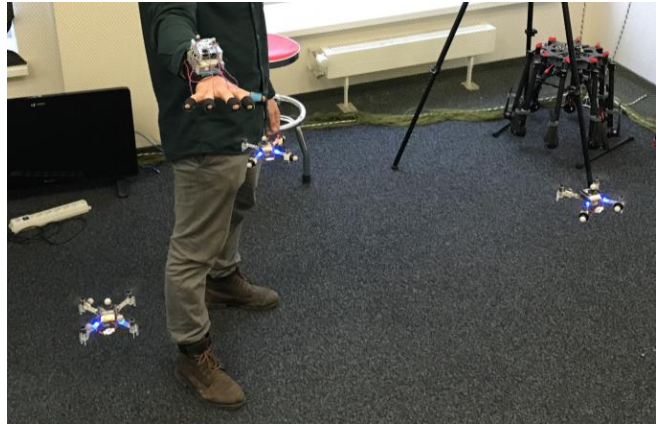

Fig. 1. Human manipulates the formation of drones.

## I. INTRODUCTION

Navigation of the quadcopters with a remote controller is a challenging task for most of the users. In order to operate a single drone in an intuitive manner, the hand commands were proposed in [1]. However, it is known that a group of robots could perform much better than a single robot due to its scalability and robustness [2]. In accordance with that, swarm robotics has many diverse applications, especially in cases when a big amount of data has to be gathered in a short period of time or when robots have to cooperate, to achieve a common goal.

For many kinds of missions, autonomous formation flight is suitable. However, for some specific applications, fully or partially guided groups of robots is the only possible solution. The operation of swarm represents a significantly more complicated task as a human has to supervise several agents simultaneously. In order for the human to work with the drone formation side by side, robust and natural interaction technics have to be developed and implemented. Human-swarm interaction (HSI) combines many research topics, which are well described by authors in [3], and could vary from communication channels to a level of swarm autonomy. Here, we focus on the interface (control and feedback) between a human operator (leader) and a swarm of robots, addressing nascent and dynamic field of HSI.

All authors are with the Intelligent Space Robotics Laboratory, Skolkovo Institute of Science and Technology, Moscow, Russian Federation.
Email: {Evgeny.Tsykunov, Luiza.Labazanova, Akerke.Tleugazy, D.Tsetserukou}@skoltech.ru

For the cases when the human is considered as a leader, standard control techniques have been developed in the last decades. Applications could include single robot-human interaction and multi robot-human interaction, both in the framework of centralized and decentralized architectures. A survey [4] shows some of the control approaches. To make human-swarm and human-environment interaction natural and safe, we have developed impedance interlinks between the drones. In contrast to the traditional impedance control [5], we propose to calculate the external force, applied to the virtual mass of impedance model, in such a way that it is proportional to the human hand velocity. The impedance model generates the desirable trajectory which reacts to the human arm motion in a compliant manner, avoiding rapid acceleration and deceleration.

Changes in the current state of the formation have to be estimated by a human operator. The importance of this statement increases with the number of robots. Haptic feedback can improve the awareness of drone formation state, as reported in [6], [7], [8], and [9]. S. Scheggi et al. [10] proposed the haptic bracelet with vibrotactile feedback to inform an operator about a feasible way to guide a group of mobile robots in terms of motion constraints. Wearable arm-worn tactile display for presentation of the collision of a single flying robot with walls was proposed in [11]. Vibrotactile signals improved the obstacle presence. In contrast to the discussed works, this paper presents a vibrotactile glove for interaction of human with a swarm of aerial robots by providing an intuitive mapping of the formation state to the human finger pads. We propose to deliver the tactile feedback about such state parameters of the formation that are hard to estimate from the visual feedback, i.e., formation state (extension or contraction) and state propagation direction. This information can improve the effectiveness of swarm navigation in the unstructured environment.

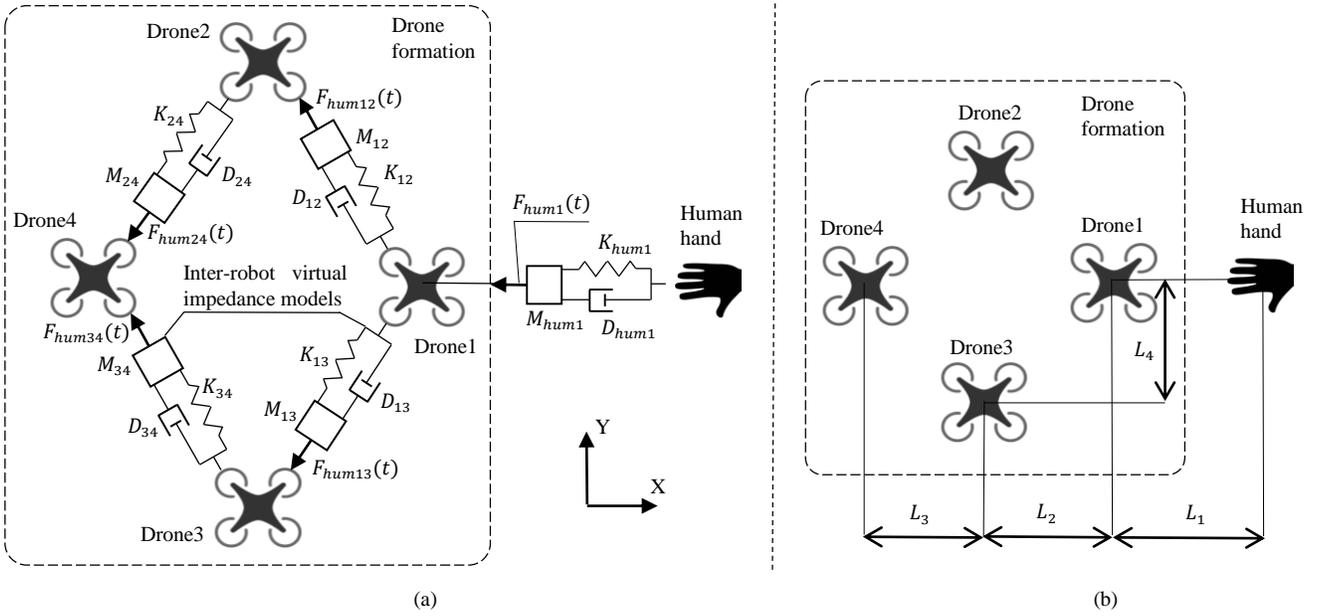

Fig. 2. (a) Position-based impedance control, (b) PID position controller. Subscription "hum" stands for human.

## II. FORMATION CONTROL

### A. Approach

To implement the adaptive manipulation of a robotic group by a human operator, such as when the inter-robot distances and formation dynamics change in accordance with the operator state, we propose a position-based impedance control [12].

In our impedance model, we introduce mass-spring-damper links between each pair of agents and between human and agent formation as shown in Fig. 2(a). The novelty of our impedance model is that we calculate the external force, applied to the virtual mass of each impedance model, in such a way that it is proportional to the operator hand velocity. The target impedance trajectory is processed by PID control securing high precision positioning, and maintaining the rhombic shape and orientation of formation (Fig. 2(b)).

While the operator is guiding the formation in space, impedance models updates the goal positions for each flying robot, which changes default drone-to-drone distances $L_i$, for ($i=1,2,3,4$). As a result, operator pushes or pulls virtual masses of inter-robot impedance models, which allows the shape and dynamics of the robotic group to be adaptive in accordance with the human hand movement. In such a way, each robot relies on the local position information coming from neighbor vehicles and, at the same time, the human operator affects all vehicles globally. We suppose that such an adaptive control could lead to a natural multi robot-human interaction. Human is capable not only to control the swarm shape in dynamics but also generate a life-like behaviors of agents which is inherent to, e.g., honeybee swarm.

### B. Math

In order to calculate the impedance correction term for the robot's goal positions, we have to solve a second-order differential equation (1) that represents the impedance model. For moving in three-dimensional space, we have to solve three differential equations for every drone, one for every axis:

$$M_d \Delta \ddot{x} + D_d \Delta \dot{x} + K_d \Delta x = F_{ext}(t) \qquad (1)$$

where $M_d$ is the desired mass of virtual body, $D_d$ is the desired damping, and $K_d$ is the desired stiffness, $\Delta x$ is the difference between the current $x_{imp}^c$ and desired $x_{imp}^d$ position $x_{imp}^c - x_{imp}^d$, and $F_{ext}(t)$ is an external force, applied to the mass. By selecting the desired dynamics parameters for the impedance model we can get various behavior.

In discrete time-space, after integration, we could write the impedance equation in the following way:

$$\begin{vmatrix} \Delta x_{k+1} \\ \Delta \dot{x}_{k+1} \end{vmatrix} = A_d \begin{vmatrix} \Delta x_k \\ \Delta \dot{x}_k \end{vmatrix} + B_d F_{ext}^k \qquad (2)$$

where $A_d = e^{AT}$, $B_d = (e^{AT} - I)A^{-1}B$, $T$ is the sampling time, $I$ is the identity matrix, and $e^{AT}$ is the state transition matrix.

Impedance model, as the second order model, could be classified by the shape of the step response. It is known that to have critically damped response, the eigenvalues of the matrix $A$ have to be equal, real, and positive. The most challenging part in here is to compute the term $e^{AT}$. Matrix exponential is refined form Cayley-Hamilton theorem, according to which every matrix satisfies its own characteristic polynomial. Using this statement, we can find:

$$A_d = e^{\lambda T} \begin{vmatrix} (1 - \lambda T) & T \\ -bT & (1 - \lambda T - aT) \end{vmatrix}$$

$$B_d = -\frac{c}{b} \begin{vmatrix} e^{\lambda T}(1 - \lambda T) - 1 \\ -bT e^{\lambda T} \end{vmatrix}$$

where $a = -\frac{D_d}{M_d}, b = -\frac{K_d}{M_d}, c = \frac{1}{M_d}$. $A_d$ and $B_d$ matrixes could be used for calculating of current $x_{imp}^c$ position of the impedance model using equation (2).

As far as we want the human operator could change the formation shape and dynamics while moving in space, then the external force term $F_{ext}(t)$ could be the function of some human state parameter. As soon as human operator guides the group of robots with a glove, we propose to calculate the external force as the function of human hand velocity:

$$F_{ext}(t) = K_v v(t) \quad (3)$$

where $K_v$ is a scaling coefficient, which determines the effect of the human operator velocity $v(t)$ on the formation.

Described above method is used to calculate impedance correction term $x_{imp}$ or the current position of virtual body of each impedance model. In order to demonstrate performance under assumption on the boundedness of the external inputs, impedance term $x_{imp}$ value is limited with the maximum value:

$$x_{imp} \leq x_{imp\_limit} \quad (4)$$

where $x_{imp\_limit}$ is the safety threshold that prevent an overrun of the impedance term.

Finally, goal positions along X-axis for each quadrotor determined in the following way (for the control structure presented in Fig.2(b)):

$$\begin{bmatrix} x_{1\_goal} \\ x_{2\_goal} \\ x_{3\_goal} \\ x_{4\_goal} \end{bmatrix} = \begin{bmatrix} x_{human} - L_1 - x_{imp\_hum1} \\ x_1 - L_2 + x_{imp\_12} \\ x_1 - L_2 + x_{imp\_13} \\ \frac{x_2 + x_2}{2} - L_3 + \frac{(x_{imp\_24} + x_{imp\_34})}{2} \end{bmatrix} \quad (5)$$

where $x_{imp\_ij}$ for (i,j=hum,1,2,3,4) are corresponding impedance correction terms, $L_i$ for (i=1,2,3,4) are base displacement for the quadrotors, as could be seen in Fig. 2(a).

## C. Verification

We use the formation of four Crazyflie 2.0 quadrotors to perform the verification flight tests. The Crazyflie 2.0 quadrotor is one of the smallest commercially available drones, which is used in many applications, such as [13]. Firstly, small size (9 cm$^2$) and weight (27 grams) provide safety, which is required for applications that involve human participation. Although we have not experienced collisions with humans, we followed simple safety rules, e.g., wearing safety glasses and gloves. Secondly, small size leads to small inertia parameters, that helps to react to control inputs promptly enough. To track the quadrotors and human, we use OptiTrack motion capture system with 8 cameras covering 5 m × 5 m × 5 m space. We used the Robot Operating System (ROS) Kinetic framework to run the development software and ROS stack [14] for Crazyflie 2.0. For communication, we applied a 2.4 GHz Crazyradio. Position and attitude update rate is 80 Hz for all drones.

As a preliminary experiment, the selection of the impedance parameters has been carried out. All PID coefficients for attitude and position controller set to default values for Crazyflie 2.0, according to [14]. Firstly, desired mass $M_d$, desired damping $D_d$, desired stiffness $K_d$ of the impedance model have been set in order to get a critically damped response which would be smooth and comfortable for a human ($M_d = 1.9, D_d = 12.6, K_d = 21.0$). Secondly, to get the safe but noticeable amplitude of the response, human velocity coefficient $K_v$, used for force calculation as in (3), is selected. We assume that the impedance correction of the goal position has to be no more than 30-50% from the distance to the neighbors $L_i$ (which is 0.5 meters in our case). We also estimated that the normal human hand velocity does not go over 1.5 m/sec while manipulating the formation. Based on this information we selected $K_v$ to be of -7 N·sec/m. A negative $K_v$ value is used because when the human is moving in one direction, drones retreat towards the opposite direction. The faster human is moving, the more lag in distance between a human and a drone. This kind of behavior is natural and safe, and could be noticed on the highway when distances between the cars naturally increase with velocity. Finally, for the safety reasons, we set the limit of impedance correction term $x_{imp\_limit}$ to be 0.25 meters for the experiments. For the simplification purposes, we use the same dynamic parameters for all impedance models in the experiment.

After the selection of all of the impedance parameters, we had to check the single drone behavior, while being guided by the human operator with proposed impedance controller. For that, we took Drone 1 and the human wearing a glove, as seen in Fig. 2. We present the values along Y-axis. Human hand velocity and impedance correction term are shown in Fig. 3. From Fig. 3 it is possible to notice that the impedance model changes its state smoothly in accordance with human hand movement. Due to the negative velocity coefficient $K_v$, human and impedance model are moving in the opposite directions. It is also possible to notice (for time range 8.5 – 9 seconds in Fig. 3), that the safety threshold $x_{imp\_limit}$ helps to prevent dangerous behavior due to high values of the input parameter (human velocity).

Fig. 4 shows the actual position of human hand along with goal and actual Drone 1 position. According to Fig. 2(b) y-coordinates of the Drone 1 and the human have to be equal, in the case of simple PID controller, and therefore human actual position and drone goal position along *Y*-axis would match each other exactly. However, due to the impedance correction of the goal position, in Fig. 4 we could notice that the drone goal position is slightly behind the human position (this difference equals to impedance term). The result could be represented as a sort of filtering of the robot goal position, which leads to the smoother drone guidance, especially in the case of extreme external inputs. It is possible to notice a delay between a human command and a drone reaction, which is obvious due to the nature of impedance controller. Fig. 4 (orange and gray lines) demonstrates that a drone position controller could track the desired trajectory accurately.

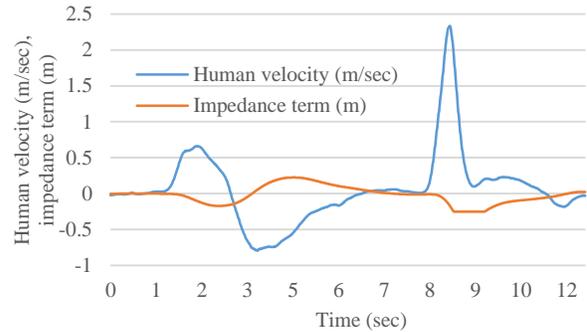

Fig. 3. Human hand velocity (blue) and impedance correction term (orange) versus time.

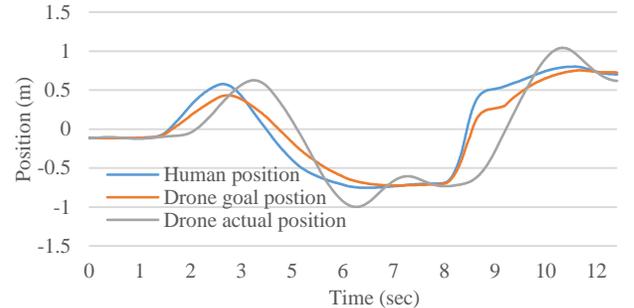

Fig. 4. Human hand position while guiding the drone (blue), drone goal position while following the human (orange), drone actual position (gray) versus time.

To demonstrate the performance of the proposed algorithm for the formation guidance, first of all, we refer to Fig. 5, where a human moves their hand from the right to the left. This figure presents an interesting feature of the impedance control. While the human starts to move fast, the formation immediately spreads along human movement direction like a spring. This example demonstrates that the formation adapts to human, right after the operator starts to move its hand. When the human hand velocity starts to decrease, the formation contracts back to the initial shape. The axis, along which the formation changes its shape, coincides with the human velocity vector; therefore, when the human changes the direction of movement, the formation adapts accordingly. Fig. 6 shows distance along the *Y*-axis between Drone 1 and Drone 4, which are placed in accordance with Fig. 2. The displacement between drones is displayed in Fig. 7. Here one can see that the magnitude of the displacement is increasing for drones farther away from the human. To meet that issue, in the future work, the stability of the proposed method will be considered with the increased number of drones.

The proposed control algorithm could be used not only for human-swarm interaction (HSI) but also for obstacle/collision avoidance. On the other hand, HSI could significantly benefit if we couple described control method with tactile feedback, forming an interface (control and feedback) between a human and a formation. Informing a human operator about the dynamic formation state (extension or contraction) at the current time could potentially improve controllability.

### III. SwarmGlove: Vibrotactile Wearable Glove

#### A. Technology

The wearable tactile displays, e.g., LinkTouch can represent multimodal information at the fingertips, i.e. force vector, vibration, and contact state [15]. However, vibration motors, being simple in control, are widely applied in Virtual Reality [16], [17]. We applied ERM vibration motors which deliver the swarm dynamic state in the form of tactile patterns.

We have designed a prototype of the tactile display with five ERM vibrotactile actuators attached to the fingertips as shown in Fig. 8(a). The vibration motors receive control signals from an Arduino UNO controller. The unit with Arduino UNO and battery are worn on a wrist as a portable device. Infrared reflective markers are located on the top of the unit. Frequency of vibration motors are changed according to the applied voltage. From Fig. 8(b), the haptic device diagram can be observed. The glove microcontroller receives values of

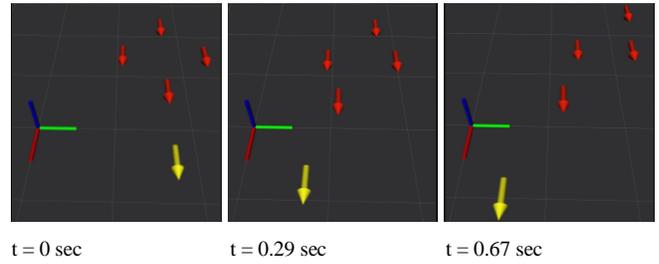

t = 0 sec  t = 0.29 sec  t = 0.67 sec

Fig. 5. Formation of four quadrotors following a human. Yellow arrow represents the human actual position and red arrows represent quadrotors goal positions. Human moves from the right to the left side.

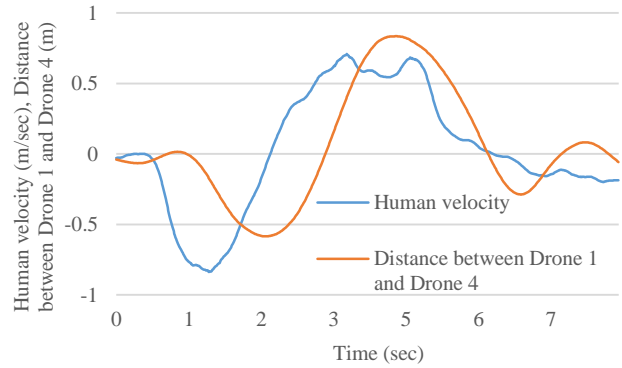

Fig. 6. Blue line represents human hand velocity versus time, orange line represents distance between Drone 1 and Drone 4.

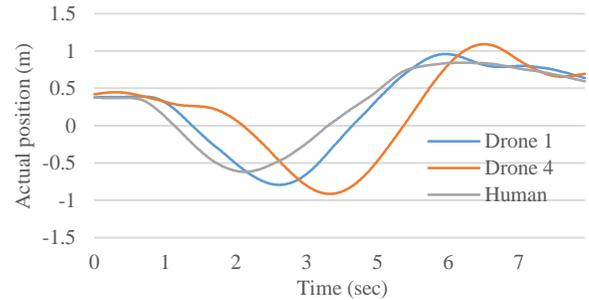

Fig. 7. Actual positions of Drone 1 (blue), Drone 4 (orange), human hand (gray) versus time.

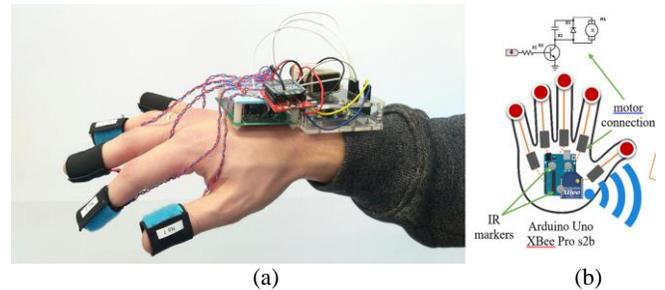

(a)  (b)

Fig. 8. (a) - wearable haptic display, (b) - haptic device diagram.

the formation state parameters from the PC. The Bluetooth and USB communications between the computer and haptic device presented in previous studies [10] and [15] are limited in working distance and mobility. Therefore, we implemented radio frequency connection through XBee Pro s2b radio modules due to its robustness and high speed of data exchange. After the Arduino UNO gets the information about the current swarm state, it applies an appropriate vibration pattern.

## B. Tactile patterns

We designed the tactile patterns for presenting the feeling of the swarm behavior at the operator's fingertips. Our motivation on the selection of the particular tactile pattern was to bring the valuable information that potentially can improve the quality (speed, safety, precision) of operation of the swarm. Fig. 9 shows the information to be presented with the wearable tactile interface, i.e., inter-robot distance and whether it is increasing or decreasing.

During swarm manipulation by the operator, the formation can change its shape, becoming contracted or extended. Therefore, the operator should take this information into account, since it contributes to better swarm operation in a cluttered environment. For instance, if the swarm gets too contracted, there is a risk of a collision between the drones. On the other hand, while guiding the formation through the obstacles, the extended state of the swarm can also lead to the collision. The distance between drones is presented by the tactile flow propagation. If the formation is extended, the flow goes from the middle finger to the side ones (Fig. 10 (a, b, c)), otherwise, the flow goes from the side fingers to the middle one (Fig. 10 (d, e, f)).

However, in many cases, the formation state is changing dynamically. In such a scenario, additional real-time information on state propagation direction could be provided to the human operator, in particular, whether the drones are flying away from each other (distance between agents is increasing) or the drones are flying toward each other (distance is decreasing). The change of distance is presented by the gradient of the tactor vibration intensity, e.g., if the distance is increasing, then side vibration motors have a higher intensity than the middle one, see Fig. 10 (a, d).

### A. Experiments and Results

#### 1) Participants

Six right-handed volunteers (3 males and 3 females, aged 22-28) participated in the experiment. They were given a period for training (10-15 minutes) so that they could get used to the sensations and learn to recognize the signals. All participants positively responded to the device convenience and level of perception.

#### 2) Experiment conditions

Optimal sensitivity of the skin is achieved at frequencies between 150 and 300 Hz [18]. Therefore, for 3 vibration levels, we assigned average frequency values: 150 Hz, 200 Hz, 250 Hz (refer to three grayscale colors shown in Fig. 3). Tactile pulses lasted for 200 and 300 ms depending on the pattern, since distinguishing tactile patterns is easier with stimulus duration in the range of 80 to 320 ms [18].

#### 1) Detection of multi-modal patterns

The experiment was devoted to the detection of multi-modal patterns. The state of the formation was mapped by the vector of propagation of tactile stimuli (e.g. if the swarm is extended firstly the third finger is activated, then the second and fourth fingers, and finally the first and the fifth ones, see Fig. 10(a) for reference). The change of distance between drones was modulated by the gradient of the vibration intensity. To emphasize the direction of the gradient, we introduced different durations of the tactile stimulus. The duration of the tactile pulse in the case of low (150 Hz) intensity was 200 ms,

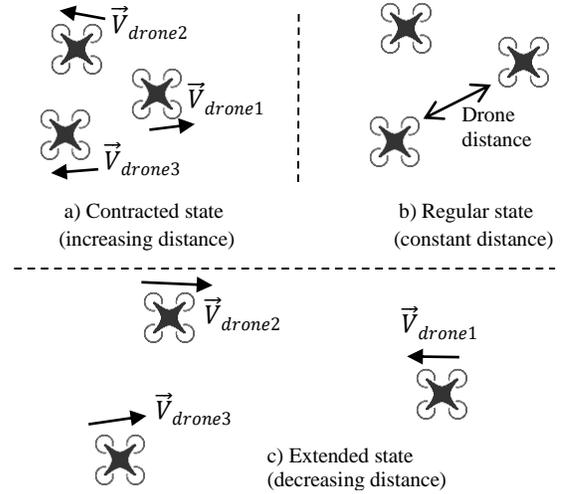

a) Contracted state (increasing distance)
b) Regular state (constant distance)
c) Extended state (decreasing distance)

Fig. 9. Contracted, regular and extended state of the formation.

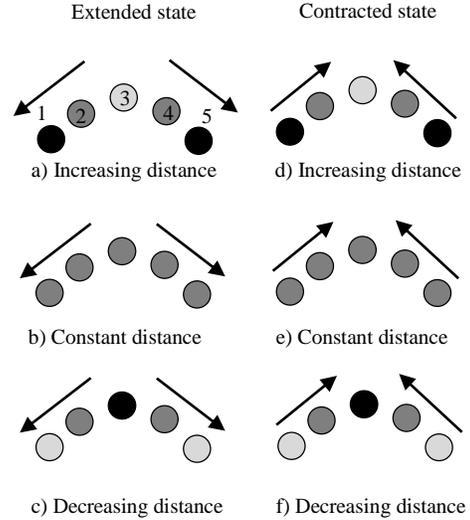

Extended state — Contracted state

a) Increasing distance — d) Increasing distance
b) Constant distance — e) Constant distance
c) Decreasing distance — f) Decreasing distance

Fig. 10. Tactile patterns for representing the state of the formation in terms of drone-to-drone distance. Each circle represents finger of a right hand (view from the dorsal side of the hand). The gray scale color represents the intensity of tactor vibration.

meanwhile, the duration of the tactile pulses with the middle (200 Hz) and high (250 Hz) intensities was 300 ms. The time interval between each stimulus wave was 600 ms. Each of the subjects experienced 60 stimuli (6 patterns were repeated 10 times in a random order).

The results of the user study for the experiment are listed in Table 1. The diagonal term of the confusion matrix indicates the percentage of the correct responses of participants. The first letter in the abbreviation of the pattern name stands for the current formation state (contraction or extension), while the second letter stands for the state propagation direction (how the drone-to-drone distance is changing: increasing, constant, or decreasing).

In order to evaluate the statistical significance of the differences between patterns, we analyzed the results of the user study using single factor repeated-measures ANOVA, with a chosen significance level of $p<0.05$.

TABLE 1. GROUP PERCENTAGE RECOGNITION OF PATTERNS FOR EXPERIMENT

| Percentage, % | Subject Response | | | | | |
|---|---|---|---|---|---|---|
| Actual pattern | CD | CI | CC | ED | EI | EC |
| Contracted state, Decreasing distance (CD) | **98.3** | 0 | 0 | 0 | 1.7 | 0 |
| Contracted state, Increasing distance (CI) | 3.3 | **86.7** | 8.3 | 1.7 | 0 | 0 |
| Contracted state, Constant distance (CC) | 10.0 | 5.0 | **85.0** | 0 | 0 | 0 |
| Extended state, Decreasing distance (ED) | 1.7 | 0 | 0 | **86.7** | 0 | 11.7 |
| Extended state, Increasing distance (EI) | 3.3 | 3.3 | 0 | 8.3 | **66.7** | 18.3 |
| Extended state, Constant distance (EC) | 0 | 0 | 0 | 3.3 | 3.3 | **93.3** |

The results of the experiment revealed that the mean percent correct scores for each subject averaged over all six patterns ranged from 78.3 to 96.7 percent, with an overall group mean of 86.1 percent of correct answers. Table 1 shows that the distinctive patterns CD and EC have higher percentages of recognition 98.3 and 93.3, respectively. On the other hand, patterns CC and EI have lower recognition rates of 85 and 66.7 percent, respectively. For most participants, it was difficult to recognize pattern EI, which usually was confused with pattern EC. Therefore, it is required to design more distinctive tactile stimuli to improve the recognition rate in some cases.

The ANOVA results showed a statistically significant difference in the recognition of different patterns ($F(5, 30) = 3.09$, $p = 0.023 < 0.05$). The paired t-tests showed statistically significant differences between the EI and EC ($p=0.015<0.05$), CD and CI ($p=0.017<0.05$), CD and CC ($p=0.0007<0.05$), CD and EI ($p=0.019<0.05$). However, the results of paired t-tests between other patterns did not reveal any significant differences, thus these patterns have nearly the same recognition rate.

IV. CONCLUSION

We have proposed a novel system SwarmTouch which integrates impedance control and tactile glove for intuitive and effective swarm control by an operator. The impedance links between agents allow the swarm to not only generate safe trajectory but also to exhibit a life-like behavior. We also designed the tactile patterns for glove and conducted experiments to reveal more distinguishable ones. The possible application of the proposed system is the navigation of swarm in the city with skyscrapers and for rescue operations.